%% file: main.tex
\relax
\documentclass[letterpaper]{llncs} %
\usepackage{url}  %
\usepackage{graphicx}  %
\usepackage[english]{babel}
\usepackage{blindtext}
\usepackage{multirow}
\usepackage{booktabs}
\usepackage{soul}
\usepackage{mathtools}
\usepackage{amsmath}
\usepackage{amssymb}
\usepackage{todonotes} %
\usepackage{xcolor}
\usepackage{color}
\usepackage{tabularx}
\usepackage{makecell}

\usepackage{enumitem}

\usepackage{grffile}	%
\usepackage[caption = false]{subfig}

\usepackage{amssymb}

\newcommand{\ru}[1]{{\color{red}{\underline{#1}}}}
\newcommand{\norm}[1]{\left\lVert#1\right\rVert}

\newcommand{\groups}{\textit{groups}}
\newcommand{\group}{\textit{group}}
\newcommand{\Groups}{\textit{Groups}}
\newcommand{\SEQL}{\texttt{SEQL}}
\newcommand{\EmbSEQL}{Emb-\texttt{SEQL}}

\newcommand{\R}{\mathbb{R}}

\begin{document}

\title{Background Knowledge Injection for Interpretable Sequence Classification}
\author{Severin Gsponer\inst{1}
\and Luca Costabello\inst{2}
\and Chan Le Van\inst{2}
\and Sumit Pai\inst{2}
\and Christophe Gueret\inst{2}
\and Georgiana Ifrim\inst{1}
\and Freddy Lecue\inst{3}}
\institute{Insight Centre for Data Analytics, University College Dublin, Dublin, Ireland, \email{\{firstname.lastname\}@insight-centre.org} \and
Accenture Labs, Dublin, Ireland, \email{\{firstname.lastname\}@accenture.com} \and
Inria, Sophia Antipolis, France, \email{freddy.lecue@inria.fr}}

\maketitle

\begin{abstract}
Sequence classification is the supervised learning task of building models that predict class labels of unseen sequences of symbols. 
Although accuracy is paramount, in certain scenarios interpretability is a must. 
Unfortunately, such trade-off is often hard to achieve since we lack human-independent interpretability metrics.
We introduce a novel sequence learning algorithm, that combines (i) linear classifiers - which are known to strike a good balance between predictive power and interpretability, and (ii) background knowledge embeddings. 
We extend the classic subsequence feature space with groups of symbols which are generated by background knowledge injected via word or graph embeddings, and use this new feature space to learn a linear classifier.
We also present a new measure to evaluate the interpretability of a set of symbolic features based on the symbol embeddings. 
Experiments on human activity recognition from wearables and  amino acid sequence classification show that our classification approach preserves predictive power, while delivering more interpretable models. 

\end{abstract}

\blindmathtrue

\input{introduction.tex}

\input{sota.tex}

\input{background.tex}

\input{proposal.tex}

\input{experiments.tex}

\input{conclusion.tex}

\bibliographystyle{splncs04.bst}
\bibliography{references}

\end{document}

%% file: introduction.tex
\section{Introduction}\label{sec:intro}

Sequence classification - the task of assigning classes to sequences of atomic symbols - occurs in a multitude of applicative scenarios such as ubiquitous computing, bioinformatics, finance, and security surveillance~\cite{DBLP:journals/sigkdd/XingPK10}.
A concrete example is the determination of protein types base solely on the amino-acid sequence. 
Deep neural architectures deliver excellent predictive power, at the expense of human interpretability~\cite{doshi2017towards} and high demand on computational resources.
Other sequence classification approaches provide better interpretability (e.g. linear models), but this is often achieved at the expense of predictive power. Not only is such an accuracy-interpretability trade-off hard to achieve, but comparing models interpretability is often left to manual inspection, as there are no agreed-upon, human-independent measures available.
At the same time, a large number of textual and graph-based background knowledge bases are available on the web. Nevertheless, there are no existing works that merge predictive models trained on sequences of symbols with prior knowledge from external knowledge bases.

\begin{figure}[htb]
  \centering
 \vspace{-1em} 
  \includegraphics[width=0.85\columnwidth]{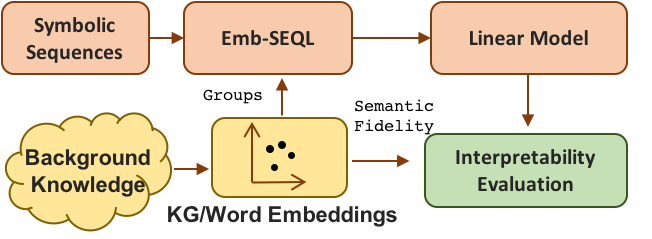}
  \vspace{-1em}
  \caption{Simplified overview of our contribution: Injecting domain knowledge for interpretable sequence classification.}
  \label{fig:overview}
\end{figure}

In this work, we focus on the problem of designing an interpretable model for sequence classification. The problem consists of two parts: i) conceiving the model itself, and ii) validating the improvement in interpretability with a proper metric.
Unlike existing sequence classification models, the intuition behind our work is that auxiliary, external background information can i) enhance the interpretability of sequence classification models, and ii) help measure such interpretability.  
We show that linear models for classification - which are known to strike a good balance between predictive power and interpretability - can be enriched with auxiliary background knowledge to obtain a quantifiable improvement in the interpretability of their features, without affecting the predictive power. Our contribution (Figure~\ref{fig:overview}) includes:

\begin{itemize}  
\item \textbf{Emb-SEQL}: a feature selection and learning algorithm for sequence classification that uses external embeddings to refine the selection of candidate features. %
\item \textbf{Semantic Fidelity}: a metric to quantify the interpretability of features extracted from symbolic sequences. The metric casts the problem into computing distances in a background knowledge embedding space, and does not depend on human-grounded evaluation protocols.

\end{itemize}	%

\vspace{-0.2cm}
\vspace{-0.2cm}

We evaluate our approach on human activity recognition from wearables (HAR) and amino acid sequence classification. We assess both predictive power and interpretability, experimenting with pre-trained word embeddings and knowledge graph embeddings. We find that using auxiliary knowledge to refine the selection of candidate features results in more interpretable models. At the same time, it does not reduce the predictive power of the learned model.  
Links to data and code will be inserted in the final version of the paper.

%% file: sota.tex
\section{Related Work}\label{sec:relatedwork}
\textbf{Sequence Classification:}
Learning classification models for symbolic sequences often uses the presence or the frequency of consecutive groups of \(k\) symbols, so called \(k\)-mers (or \(n\)-grams in text processing) as features~\cite{DBLP:journals/sigkdd/XingPK10}.
Support Vector Machines (SVMs) show promising results: specific string kernels have been proposed, as well as implementation tricks that improve their efficiency~\cite{Leslie2002,Sonnenburg2005,Sonnenburg07Large}.
Markov Models and Hidden Markov Models~\cite{Rabiner1989} model the probability distribution of sequences for each class separately and assign the class with the highest likelihood to unseen sequences at inference time.
Current state-of-the-art results are held by Convolutional Neural Networks (CNN) that operate on sequences of characters, which have been successfully applied to sequences~\cite{Zhou2015,Alipanahi2015}. %
Nevertheless, CNNs and SVMs are black box models and have poor interpretability. %

\noindent\textbf{Background Knowledge Injection:}
Background knowledge is typically used to improve accuracy: auxiliary knowledge can be encoded as rules, for more accurate relation extraction~\cite{DBLP:conf/naacl/RocktaschelSR15}, or to predict missing links in knowledge graphs~\cite{DBLP:conf/uai/MinerviniDRR17}.
There have been attempts to incorporate semantic monotonic constraints derived from background knowledge~\cite{freitas2014comprehensible}, but not for sequence classification.

\noindent\textbf{Interpretability Metrics:} 
\cite{doshi2017towards} discuss evaluation protocols to assess the interpretability of machine learning models.
They take into account human-grounded experiments - with real-world and simplified tasks.
Besides, they also acknowledge the need for \textit{functionally-grounded} protocols that replace human intervention with proxy tasks.
A simple proxy to compare linear classifiers is measuring the size of the model (number of features with non zero weights) - the assumption being that the smaller the model, the higher the interpretability.
Nevertheless, this is an over-simplistic assumption, as size does not capture the semantics of the model features~\cite{freitas2014comprehensible}.

%% file: background.tex
\section{Preliminaries}\label{sec:background} 
\subsection{Sequence Classification}
Learning a mapping from sequences of symbols to categorical labels is commonly known as sequence classification.
Let $D = \{(s_1, y_1), (s_2, y_2), \dots, (s_N, y_N)\}$ be a sequence database of instance-label pairs, where \(s_i\) is a sequence and \(y_i \in L\) the corresponding label.
The goal of sequence classification is to learn a mapping \(\xi\) from the sequence database \(D\) so that we can predict the label of a yet unlabeled sequence \(s\in S\).
Formally such a mapping is a function \(\xi: S\rightarrow L\) where \(S\) is the set of all possible sequences and \(L=\{c_1, c_2, \dots, c_N\}\) the set of class labels.
A sequence \(s_i \in S\) has the following form \(s_i=\langle \sigma_{i1},\sigma_{i2},\dots,\sigma_{in_{i}} \rangle\) and each of the individual symbols \(\sigma_{ij}\) belongs to a predefined finite alphabet $\Sigma$.
For example, if \(\Sigma = \{\text{A}, \text{B}, \text{C}\}\) a sequence could be \(s_1 =\langle \text{B},\text{A},\text{B},\text{C} \rangle \).
Note that the lengths \(n_i\) of sequences is variable.

A \(k\)-mer is a sequence of \(k\) consecutive symbols, e.g., \(k'=\langle \text{B}, \text{A},\text{B}\rangle\).
We write \(k' \subseteq s_{i}\) and say \(k'\) is present in \(s_{i}\) if an exact match of \(k'\) is found in \(s_{i}\).
Given this definition and an enumeration schema of all \(k\)-mers present in the training data, we can represent a sequence $s_i$ as a binary vector:
$\mathbf{x}_i = (x_{i1},\dots, x_{ij},\dots, x_{in})^{T}, x_{ij} \in \{0,1\}, i= 1, \dots, N$, where $x_{ij}=1$ means that \(k\)-mer $k_j$ occurs in sequence $s_i$.

Such a representation allows us to learn a linear model, i.e., a parameter vector $\boldsymbol{\beta}$ of feature weights to predict the class label of a sequence \(s_i\) by setting $y_i = sgn(\boldsymbol{\beta}^{T}\mathbf{x}_i)=\xi(s_i)$.
Although linear models are not powerful enough to capture non-linear relationships, by working in a very complex feature space (e.g., all $k$-mers) it is possible to learn powerful models, similar to the kernel trick applied by kernel Support Vector Machines \cite{Ifrim2011}.

\subsection{SEQL}
Although the entire \textit{\(k\)-mer space} is huge and in practice infeasible to generate explicitly, it can still be used by exploiting the nested structure of the feature-space using \SEQL~\cite{Ifrim2011}. 
In this work we adopt \SEQL\, a linear sequence classifier algorithm, as we want to learn a model that is interpretable but still achieves high accuracy.
The main idea behind \SEQL\  is to use a greedy coordinate gradient descent with the Gauss-Southwell rule \cite{nutini2015cgd} which allows to avoid the explicit generation of the feature vectors~\cite{gsponer2017efficient}.
A key step of this approach is the efficient search for the current \textit{best} \(k\)-mer, in the sense of maximum absolute gradient value, followed by an update of the corresponding weight value \(\beta\).
These two steps are executed iteratively until a convergence threshold is reached.
The search part itself is realized with a branch-and-bound tree search which is made feasible by a bound on the gradient value of \(k\)-mers based on its own sub-\(k\)-mers.
In particular, each iteration starts by computing the gradient values of all \(1\)-mers whereby the best gradient value found so far is saved in \(\tau\). 
For each of the \(1\)-mer \(k'\) the corresponding upper bound \(\mu(k')\) is computed.
The sub-tree starting at \(k'\) can be pruned whenever \(\mu(k') \le \tau\) otherwise we expand \(k'\) and repeat the procedure.
This search procedure allows to find the \textit{best} \(k\)-mer in an efficient and timely manner.
The resulting model is a weighted list of \(k\)-mers which is easier to  understand by humans.
\SEQL\ has support for two classification losses i) logistic loss and ii) squared hinge loss; here, we use i) to learn linear binary sequence classification models.

Moreover, \SEQL\ goes beyond traditional \(k\)-mers, since it has the ability to use wildcards within the generated \(k\)-mers by using the \(*\)-character.
Such wildcard allows \(k\)-mers with gaps, which leads to more general features. Nevertheless this is computationally expensive~\cite{Ifrim2011}.

\subsection{Word Embeddings}

Word embeddings are representation learning techniques widely adopted in natural language processing. They map words in a text corpus to a low-dimensional, continuous vector space. Such vectors act as representations of terms in a n-dimensional metric space. %
Word embeddings are mostly generated by processing word co-occurrences, or by using neural architectures, the most popular models being word2vec~\cite{DBLP:conf/nips/MikolovSCCD13} and GloVe~\cite{pennington2014glove}.
Popular pre-trained word embeddings collections such as ConceptNet Numberbatch~\cite{speer2017conceptnet} and GloVe\footnote{\url{https://nlp.stanford.edu/projects/glove/}} are available on the web.
The main shortcoming of word embeddings is that single vectors may represent words that carry multiple meanings.

\subsection{Knowledge Graph Embedddings}
Knowledge graphs are graph-based knowledge bases whose facts are modeled as relationships between entities. Examples are DBpedia, WordNet, and YAGO. %
Formally, a knowledge graph $\mathcal{G}=\{ (sub,pred,obj)\} \subseteq \mathcal{E} \times \mathcal{R} \times  \mathcal{E}$ is a set of triples in the form $t=(sub,pred,obj)$, each including a subject $sub \in \mathcal{E}$, a predicate $pred \in \mathcal{R}$, and an object $obj \in \mathcal{E}$. $\mathcal{E}$ and $\mathcal{R}$ are the sets of all entities and relation types of $\mathcal{G}$.

Knowledge graph embedding models are neural architectures that encode concepts from a knowledge graph (i.e. entities $\mathcal{E}$ and relation types $\mathcal{R}$) into low-dimensional, continuous vectors $\in \R^k$. Such \textit{knowledge graph embeddings} have many applications, e.g., in knowledge graph completion, entity resolution, and link-based clustering~\cite{nickel2016review}.
Knowledge graph embeddings are learned by training a neural architecture over a knowledge graph. Although such architectures vary, the training phase always consists in minimizing a loss function (usually negative log-likelihood or hinge loss) $\mathcal{L}$ that includes a \textit{scoring function} $f_{m}(t)$, i.e., a model-specific function that assigns a score to a triple $t$. 
The optimization procedure learns optimal embeddings by minimizing $\mathcal{L}$, such that the model assigns high scores to true statements, and low scores to statements unlikely to be true.

%% file: proposal.tex
\section{Method}\label{sec:method}
In this section we describe \EmbSEQL, our background knowledge-enriched sequence classification model. We also present the Semantic Fidelity, the metric that we use to assess the interpretability of the features learned by \EmbSEQL.

\subsection{Emb-SEQL}
One of the drawbacks of the \(k\)-mer-based approach of \SEQL\ is that it fully relies on matching exact \(k\)-mers.
The \(*\)-wildcard relaxes this constraint but it is very general as it allows an arbitrary symbol.
As an alternative approach, we introduce the concept of \groups.\
The main intuition behind these \groups\ is that there exist symbols in the alphabet that are exchangeable in certain situations.
Conceptually, \groups\ form a new symbol that can be considered as \texttt{OR} combination of multiple base symbols from the original alphabet.
We use these new symbols to extend the all-\(k\)-mer representation that \SEQL\ uses and write them, similar to a regular expressions, as \((\text{A}|\text{B})\).
\Groups\ can be formed by hand but we are more interested in forming them automatically by exploiting background knowledge.

Symbols in more complex alphabets (e.g., Activities, NLP) often have relationships between each other in the sense that some symbols are semantically closer than others.
We use the embedded representations of symbols to measure such closeness and form \groups\ based on this measurement.
To find sensible \groups\ automatically we first map all base symbols into the embedding space followed by clustering.
Various (overlapping) clustering techniques can be used for this task.
We adopt a simple radius-based approach: for each symbol \(\sigma\) in the alphabet a \group\ is formed by aggregating all the symbols that fall within a fixed radius $r \in \R$ around the embedding of the symbol \(\sigma\). 
After collecting the \groups\ around each individual symbol, all exact duplicates are removed to obtain a final list of \groups.
It is clear that the selection of radius \(r\) is crucial, as it directly determines the \group\ sizes.
If chosen too small, no \groups\ are formed; on the other hand, a radius too big leads to large and general \groups\ and eventually to only one \group\ that contains all symbols (and hence emulates the \(*\)-wildcard).
Currently, we rely on manual selection of an appropriate radius. Our initial tests with a $k$-nearest neighbourhood approach as an alternative to the radius-based approach did not achieve better performance. Further  exploration of automatic group selection mechanisms is left for future work. 

We extend \SEQL\ by first pre-computing \groups\ followed by the normal \SEQL\ learning procedure. We call this \EmbSEQL\ for \textbf{Emb}edding enriched \textbf{SEQL}. 
Once the \groups\ are generated, each of them acts as a new base symbol for \SEQL\ and can be part of any \(k\)-mer.
During the tree search of \texttt{SEQL}, \groups\ behave exactly like a normal symbol of the alphabet.

\subsection{Semantic Fidelity}
Our base assumption is that binary linear classification models (\textit{i.e.,} weighted list of features) are understandable and interpretable as long as their features are. This complies with the \textit{decomposability} propriety of interpretable models proposed by \cite{Lipton:2018:MMI:3236386.3241340}. Nevertheless, determining how interpretable is a set of features is a task often neglected, and still requires manual intervention.
To overcome this problem, we propose a \textit{functionally-grounded} protocol~\cite{doshi2017towards} based on the \textit{Semantic Fidelity}, a novel metric to measure the interpretability of the features of a linear model for sequence classification without the need for user intervention. 
Following the rationale that explanations should match user expectations~\cite{miller2018explanation}, we cast the problem of measuring the interpretability of a set of features to computing distances in the embedding space of an auxiliary background knowledge base. The intuition is that features with positive weights should be highly related to the concept of target class \(c\) and in contrast negative features should relate $bar{c}$ the not-target class.

We define the Semantic Fidelity as follows:

\begin{equation}
  \label{eq.modelindex}
  SF= 1-\frac{1}{2n}  \sum_{\phi_{i} \in \Phi }h(\phi_i)
\end{equation}  

$\Phi$ is the set of features, $\phi_i \in \Phi$ is a feature, $n$ is the number of features, and $h$ is defined as: 

\begin{equation}
\label{eq.hc}
  h(\phi)=|w|\begin{dcases*}
       d(\phi,c) &{} if $w \ge 0$\\
       d(\phi,\bar{c}) & otherwise 
  \end{dcases*}
  \end{equation}

where $c$ is the positive class of the binary classification task (i.e. the target) and $\bar{c}$ the negative class, $w$ is the weight associated to feature $\phi$, and $d(\phi, c)$ is the distance between a \(k\)-mer feature $\phi$ and the concept of the target class $c$. 
The distance $d(\phi, c)$ is defined as the average distance between the embeddings $\textbf{E}_{\sigma}$ of each individual k-mer symbol $\sigma \in \phi$ and the embedding $\textbf{e}_c$ of the class:

\begin{equation}
  \label{eq.embddist}
  d_c(\phi,c) = \frac{1}{n_{\phi}}\sum_{\sigma_{j}\in \phi}\norm{\textbf{E}_{\sigma_{j}} - \textbf{e}_c} 
\end{equation}

where $n_{\phi}$ is the number of symbols in $\phi$ and the embedding $\textbf{E}_{\sigma}$ of symbol $\sigma$ represents a single symbol, or in case of an \EmbSEQL\ \group\ of length $n_{\sigma}$, the average of all symbols $\tau$ in the group:

\begin{equation}
\label{eq.dsdsd}
  \textbf{E}_{\sigma}=
  \begin{dcases*}
       \textbf{e}_{\sigma} & if $n_{\sigma} = 1$ \\
       \frac{1}{n_{\sigma}}\sum_{\tau_{k} \in \sigma}{\textbf{e}_{\tau_{k}}} & otherwise
  \end{dcases*}
  \end{equation}
  
We assume the embedding space and weights $w$ to be normalized so that ($w \in [0,1]$) and the maximum distance \(d(\phi,c)=2\), and consequently $SF \in [0,1]$, where a higher value means a more interpretable model.

%% file: experiments.tex
\section{Experiments}\label{sec:eval}
In this section we assess the interpretability and the predictive power of \EmbSEQL.
We experiment in two distinct application scenarios: human activity recognition from wearables, and amino-acid sequences classification.
In a second experiment, we show that the background knowledge injection of \EmbSEQL\ does not affect its predictive power, but improves interpretability.

\subsection{Experimental Settings}

\textbf{Datasets.}
We experiment with a number of symbolic sequence classification datasets and a range of auxiliary background knowledge sources.
The symbolic sequence datasets used in the experiments are:

\begin{itemize}[leftmargin=1cm]
\item \textbf{OPPORTUNITY (HAR)}:  
Human activity recognition dataset of wearable sensor data collected from subjects performing actions in a room~\cite{roggen2010}.
It includes inertial measurements from 15 subjects, resulting in 113 sensor recordings provided as multivariate time series.
Data points are annotated at different levels of abstraction.
For this paper, we aggregate the four low-level labels (\textit{left hand action, left hand object, right hand action, right hand object}) as well as the \textit{locomotion} annotations to form a 5-let.
We transform the multidimensional symbolic sequences of OPPORTUNITY by encoding 5-lets into unique symbols (we merge adjacent repeated 5-lets).
This procedure results in more than 1,400 unique symbols.
We concatenate all records for all subjects, and we window with size 1,000 (roughly 30 seconds) and stride 50.
We label a window with its majority class. 
Our task is predicting the five top-level activities (\textit{Relaxing, Coffee time, Clean up, Sandwich time, Early morning}) from sequences of  \(5\)-lets.
We use a one-vs-all approach to address the multiclass setting of the dataset. All results are obtained with 10-fold cross validation.
Note that for \EmbSEQL\ we compute the embedding of a \group\ by averaging the five embeddings of symbols in a 5-let. 

\item \textbf{Protein}: An excerpt of PhosphoELM\footnote{\url{http://phospho.elm.eu.org/}} used in~\cite{zhou2016pattern}.
  It includes sequences of 21 distinct amino acids from the S/T/Y phosphorylation site. Each sequence is labeled with a protein group.
  We narrow down to two kinase groups (PKA group with 381 sequences and SRC with 157), to compare against the binary classification task results in ~\cite{zhou2016pattern}.
  We obtained the result by applying 10 fold cross validation as done in~\cite{zhou2016pattern}.
\end{itemize}

\begin{table}[ht]
  \small
	\caption{The adopted background knowledge graphs}
	\label{table:kg_stats}
	\centering
{\setlength{\tabcolsep}{1em}
	\begin{tabular}{l rrr}
		  \toprule          & WordNet & YAGO-41 & ChEBI-ChEMBL \\ %
		  \midrule
		  Triples			&  2,429,896  &  39,198,096  & 1,947,490            \\ %
		  Relations         &         36  &          41  &        46            \\ %
		  Entities          &  1,499,274  &   8,316,467  &   938,867            \\ %
		  \bottomrule
	\end{tabular}
    }
\end{table}

\noindent{}We used the following pre-trained word embeddings:
\begin{itemize}[leftmargin=1cm]
	\item \textbf{ConceptNet Numberbatch}: ConceptNet Numberbatch 17.06\footnote{\url{http://bit.ly/numberbatch}} includes word embeddings for more than 1.9M terms from the ConceptNet open data project. It combines data from ConceptNet, word2vec, GloVe, and OpenSubtitles 2016~\cite{speer2017conceptnet}. Embeddings have dimensionality $k=300$.
	\item \textbf{GloVe}: these pre-trained embeddings have been created with the GloVe unsupervised model from a large corpus of data crawled from the web~\cite{pennington2014glove}, and cover ~1.9M words. Embeddings have dimensionality $k=300$.
\end{itemize}

\noindent{}We used the following knowledge graphs (detailed statistics are reported in Table~\ref{table:kg_stats}):
\begin{itemize}[leftmargin=1cm]
	\item \textbf{WordNet}: WordNet is a popular lexical database of English terms. Words are grouped into \textit{synsets}, sets of cognitive synonyms that express a distinct concept. Synsets are connected with typed relations that represent conceptual, semantic, and lexical relations. We use the RDF version of WordNet 3.1\footnote{\url{http://wordnet-rdf.princeton.edu/about}}.
	\item \textbf{YAGO-41}: YAGO is a large, broad-scope knowledge graph. We used version 3.1\footnote{\url{http://bit.ly/YAGO3}}. Due to the large size of YAGO, we only used the following splits: \texttt{yagoDBpediaClasses}, \texttt{agoDBpediaInstances}, \texttt{yagoTaxonomy}, \texttt{yagoTypes}, and \texttt{yagoFacts}.
	\item \textbf{ChEBI-ChEMBL}: the knowledge graph includes triples from the RDF versions of ChEBI\footnote{\url{https://www.ebi.ac.uk/chebi/}} and ChEMBL\footnote{\url{http://bit.ly/ChEMBL-RDF}}. ChEBI includes information about small chemical compounds, i.e., molecular entities involved in processes of living organisms. We use the ChEBI-core split (\(\sim\)1.8M triples). ChEMBL-RDF 24.1 is a manually curated chemical database of bioactive molecules with drug-like properties. We downloaded the splits describing the target triples, and the mappings to ChEBI entities.

\end{itemize}

For each embedding, we manually select a radius for the \group\ generation. The main criteria for the selection are the total number of \groups\ as well as their size. The best radii are: GloVe: 0.35, WordNet: 0.185, YAGO-41: 0.23, ConceptNet: 0.23, ChEBI-ChEMBL: 0.65.

\textbf{Implementation Details.}
\EmbSEQL\ is implemented in C++. The Semantic Fidelity function is written in Python 3.6. 
The implementation of the knowledge graph embeddings model uses TensorFlow, on Python 3.6.

\begin{figure}[h]
  \centering
  \includegraphics[width=.8\columnwidth]{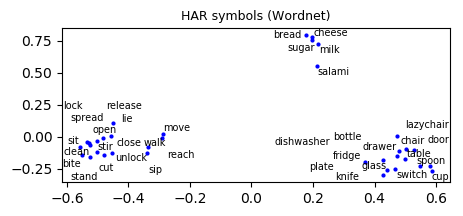}%
  \caption{PCA-reduced plot of the embeddings used in the experiments. Note the clear-cut clusters of concepts (food, kitchenware, verbs).}
  \label{fig:embeddings}
\end{figure}

\textbf{Knowledge Graph Embeddings Generation.}
Besides using pre-trained word embeddings (GloVE and ConceptNet Numberbatch), we also experiment with knowledge graph embeddings. This is done to overcome the single-vector multiple-meaning shortcoming of word embeddings. We learn knowledge graph embeddings for each knowledge graph listed in Table~\ref{table:kg_stats}. We use ComplEx~\cite{trouillon2016complex}, the neural embedding model that strikes the best trade-off between predictive power and training speed. This is crucial given the size of the knowledge graphs used in the experiments.
We rely on typical hyperparameter values known to perform well for splits of WordNet and YAGO: we train the embeddings with dimensionality $k=150$, AdaGrad optimizer, initial learning rate $\alpha_0=0.1$, margin-based pairwise loss function with margin $\gamma=2$, and negatives per positive ratio $\eta=2$, $epochs=100$.
Figure~\ref{fig:embeddings} shows a PCA-reduced scatterplot of the concept embeddings for the human activity recognition.

\subsection{Features Interpretability}

\textbf{Human Activity Recognition.}
We learn features from the OPPORTUNITY dataset with \SEQL\ and \EmbSEQL, experimenting with word and graph embeddings generated from different auxiliary knowledge bases. 
We compute the Semantic Fidelity (Equation~\ref{eq.modelindex}) of the learned features to assess if the features of \EmbSEQL\ obtained with the embedding-driven \groups\ are more interpretable than those obtained with plain \SEQL. 

\begin{figure}[htb]
  \centering
  \includegraphics[width=0.8\columnwidth]{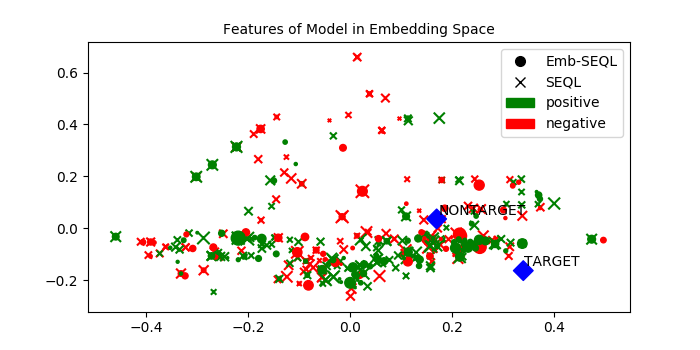}
  \caption{PCA-reduced plot of the \EmbSEQL\ and \SEQL\ models for class \textit{coffee time} in the WordNet embedding space. Note positive (negative) weighted features of \EmbSEQL\ are closer to the TARGET (NONTARGET) class concepts.}
  \label{fig:modelembedding}
\end{figure}

Table~\ref{table:relevance} reports the semantic fidelity $SF$ obtained by five binary classifiers defined for each of the five top-level activities to predict. Results are stable across the five target classes.
\EmbSEQL\ outperforms \SEQL\ with most of the embeddings: WordNet brings 4.6\% increase in Semantic Fidelity, while GloVe obtains a 2.3\% increase and ConceptNet a 0.4\% increase.
YAGO-41, on the other hand does not bring any advantage over plain \SEQL. 
This is probably due to sparse relations and lack of redundancy in the YAGO splits we used to build YAGO-41. Future experiments will use a complete version of YAGO.
Figure~\ref{fig:modelembedding} shows an example of the embedded features of \EmbSEQL\ and \SEQL\ in a PCA-reduced representation.

\textbf{Amino Acids.}
We also experiment with the Protein dataset. As for the HAR scenario, we learn features with both \SEQL\ and \EmbSEQL. For \EmbSEQL\ we use the ChEBI-ChEMBL auxiliary knowledge base, which we inject in the model as knowledge graph embeddings.
Table~\ref{table:relevance} reports the Semantic Fidelity $SF$ over a single class, as this is a binary classification task: \EmbSEQL\ reaches a Semantic Fidelity 1.5\% higher than its counterpart, thus making its features more interpretable.

\begin{table}[htb]
    \centering
    \small
    \caption{Semantic Fidelity for the human activity recognition (HAR) and Protein sequence classification tasks. Higher scores indicates more interpretable models. For the HAR experiment we report the mean semantic fidelity $\overline{SF}$ obtained by five binary classifiers defined for each of the five top-level activities to predict. We also report the results for each individual class.}
{\setlength{\tabcolsep}{0.25em}
    \begin{tabular}{@{}lrlccccccc@{}}
        \toprule

\textbf{HAR:} & &&&&&&&&\\
         Embeddings & Model & \(\overline{SF}\) & std &Class 1&Class 2 &Class 3&Class 4&Class 5\\
       \midrule 
   
                  \multirow{2}{*}{GloVe} & SEQL     & 0.902          & 0.028 & 0.930 & 0.871 & 0.925 & 0.865 & 0.921 \\
                                         & Emb-SEQL & \textbf{0.923} & 0.025 & 0.958 & 0.888 & 0.931 & 0.901 & 0.938 \\[3pt]
      
                  \multirow{2}{*}{ConceptNet}	& SEQL & 0.871 & 0.033  &0.908& 0.828& 0.895 & 0.833& 0.889\\
                 							& Emb-SEQL & \textbf{0.875} & 0.025&  0.903& 0.853& 0.887& 0.836& 0.894\\[3pt] 
      
                  \multirow{2}{*}{YAGO-41}	& SEQL & \textbf{0.867} & 0.029  & 0.899& 0.847& 0.893& 0.823& 0.872\\
                 							& Emb-SEQL & 0.835 & 0.043  &0.897& 0.824& 0.861& 0.767& 0.827\\[3pt] 
      
                  \multirow{2}{*}{WordNet}	& SEQL     & 0.894          & 0.025 & 0.921 & 0.879 & 0.918 & 0.857 & 0.895 \\
                                           & Emb-SEQL & \textbf{0.936} & 0.010 & 0.937 & 0.945 & 0.943 & 0.917 & 0.939 \\[3pt]
      \toprule
\textbf{Protein:} & &&&&&&&&\\
         Embeddings & Model & \(\overline{SF}\) \\
       \midrule 
      
                  \multirow{2}{*}{ChEBI-ChEMBL} & SEQL     & 0.708           \\
                                                & Emb-SEQL & \textbf{0.719}  \\
        						
        \bottomrule
    \end{tabular}
    }
    \label{table:relevance}
\end{table}

\subsection{Classification Quality}
\textbf{Human Activity Recognition.}
Besides interpretability, we are also interested in assessing whether \EmbSEQL\ achieves a predictive power comparable to \SEQL.
Therefore, we compare the performance of \EmbSEQL\ to \SEQL\ without any knowledge injection, as well as to a \texttt{SVM} baseline, and a LSTM-based neural network.
We use a \texttt{SVM} with RBF kernel implemented with libsvm~\cite{chang2011libsvm}.
We extracted all $k$-mers up to $k=6$ and explicitly generated the feature vector representation for each example as input for the SVM.
The LSTM has 64 hidden units followed by a single 16-unit hidden layer classifier which maps to the number of output classes. The model is trained with Adam on weighted cross entropy loss, to mitigate class imbalance. It is trained for 100 epochs with early stopping. 
\begin{table}[h]
  \centering
  \small
\caption{Results on the HAR dataset (OPPORTUNITY) and Protein dataset. Best results in bold.}
{\setlength{\tabcolsep}{1.8em}
\begin{tabularx}{\textwidth}{@{}llllr@{}}

\toprule
               Dataset   & Model                     & Embeddings   & F1             & Accuracy       \\ \midrule
\multirow{7}{*}{HAR}     & SVM                       &              & 0.502          & 0.564          \\
                         & LSTM                      &              & 0.767          & 0.810          \\
                         & SEQL                      &              & \textbf{0.973} & \textbf{0.961} \\
                         & \multirow{4}{*}{Emb-SEQL} & ConceptNet   & 0.965          & 0.951          \\
                         &                           & GloVe        & 0.961          & 0.945          \\
                         &                           & WordNet      & 0.968          & 0.955          \\
                         &                           & YAGO-41      & 0.957          & 0.941          \\ \midrule
\multirow{5}{*}{Protein} & SCIS\_MA                  &              & -              & \textbf{0.948} \\
                         & HMM                       &              & -              & 0.918          \\
                         & LSTM                      &              & 0.797          & 0.796          \\
                         & SEQL                      &              & \textbf{0.902} & 0.903          \\
                         & Emb-SEQL                  & ChEBI-ChEMBL & 0.898          & 0.901          \\ \bottomrule
\end{tabularx}
}
\label{table:results}
\end{table}

Table~\ref{table:results} shows the results of the experiment on the OPPORTUNITY dataset with the above mentioned embeddings.
We show the weighed F1 score, as well as the accuracy, excluding the null class (as done in prior work).
Results show that all \texttt{SEQL} models, regardless of the injection of auxiliary knowledge, outperform the \texttt{SVM} in both metrics, as well as the LSTM.
The performance of \EmbSEQL{} is comparable to \SEQL{}.
We conclude that the injection of knowledge into \EmbSEQL\ did not hurt the performance of the model with regard to Accuracy, 
but lead to better model interpretability (according to Semantic Fidelity).

\textbf{Amino Acids.}
A similar conclusion can be drawn for the evaluation on the Protein dataset.
Table~\ref{table:results} shows the F1 score and accuracy of \SEQL\ and \EmbSEQL\ with ChEBI-ChEMBL embeddings as well as for the LSTM-based architecture described for HAR and SCIS\_MA (sequence classification based on association rules) and its HMM baseline~\cite{zhou2016pattern}.
It is clearly visible that the Accuracy of \SEQL\ and \EmbSEQL\ lacks somewhat behind SCIS\_MA and HMM, but the injection of knowledge doesn't significantly hurt the performance of \EmbSEQL.

\iffalse
\begin{table}[t]
    \centering
    \begin{tabular}{@{\extracolsep{7pt}}llcc@{}}
        \toprule
        
         Model & Embeddings &  F1 & Accuracy
        
        \\ \hline
        
		    SVM \ru{old} &  & 0.699 & 0.705  \\
        LSTM &  & - & 0.810 \\
        SEQL &  & \textbf{0.973} & \textbf{0.961} \\

        \multirow{4}{*}{Emb-SEQL}	& ConceptNet & 0.965 & 0.951   \\
        							& GloVe & 0.961 & 0.945 \\
        							& WordNet & 0.968& 0.955 \\
        							& YAGO-41 & 0.957 & 0.941 \\
        
        \bottomrule
    \end{tabular}
    \caption{Human activity recognition on the HAR dataset (OPPORTUNITY). Best results in bold.}
    \label{table:results}
\end{table}

\begin{table}[t]
    \centering
    \begin{tabular}{@{\extracolsep{7pt}}llcc@{}}
        \toprule
        
        Model & Embeddings &  F1 & Accuracy
                                   
        \\ \hline
        
        SCIS\_MA  &  & - & \textbf{0.948} \\
        HMM  &  & - & 0.918\\
        LSTM  &  & - & \ru{??????}\\
        SEQL &  & \textbf{0.902} & 0.903 \\
        Emb-SEQL & ChEBI-ChEMBL & 0.898 & 0.901 \\
        
        \bottomrule
    \end{tabular}
    \caption{Protein groups classification task on the Protein dataset (PhosphoELM). Best results in bold. }
    \label{table:results}
\end{table}
\fi

%% file: conclusion.tex
\section{Conclusion}\label{sec:conclusion}
We show that semantic embeddings help generate more interpretable features for sequence classification with linear models.
Besides, we also show that distances in embedding spaces can be used to quantify how interpretable such features are.
Future work will include exploration of different clustering techniques to form groups in \EmbSEQL.
An important axis of work will be validating the Semantic Fidelity against human-grounded and application-grounded evaluation protocols.
Furthermore, we will investigate the application of the Semantic Fidelity to other feature-based models.